\documentclass{article}
\usepackage{eaclap}

\title{\vspace{-0.5in}Representing Text Chunks}
\author{{\rm
\begin{tabular}{cc}
{\bf Erik F. Tjong Kim Sang}&{\bf Jorn Veenstra}\\
Center for Dutch Language and Speech&Computational Linguistics\\
University of Antwerp&Tilburg University\\
Universiteitsplein 1&P.O. Box 90153\\
B-2610 Wilrijk, Belgium&5000 LE Tilburg, The Netherlands\\
erikt@uia.ac.be&veenstra@kub.nl
\end{tabular}
}}

\begin{document}

\maketitle

\begin{abstract}
Dividing sentences in chunks of words is a useful preprocessing
step for parsing, information extraction and information retrieval.
\cite{ramshaw95} have introduced a "convenient"
data representation for chunking by converting it to a tagging
task.
In this paper we will examine seven different data representations
for the problem of recognizing noun phrase chunks.
We will show that the the data representation choice has a minor	
influence on chunking performance.
However, equipped with the most suitable data representation,
our memory-based learning chunker was able to improve the best
published chunking results for a standard data set. 
\end{abstract}

\section{Introduction}

The text corpus tasks parsing, information extraction and information
retrieval can benefit from dividing sentences in chunks of words.
\cite{ramshaw95} describe an
error-driven transformation-based learning (TBL) method for 
finding NP chunks in texts.
NP chunks (or baseNPs) are non-overlapping, non-recursive noun phrases.
In their experiments they have modeled chunk recognition
as a tagging task:
words that are inside a baseNP were marked {\tt I}, 
words outside a baseNP received an {\tt O} tag and 
a special tag {\tt B} was used for the first word inside a baseNP
immediately following another baseNP. 
A text example:

\begin{quote}
original: \\
In $[_N$ early trading $_N]$ in $[_N$ Hong Kong $_N]$
$[_N$ Monday $_N]$ , $[_N$ gold $_N]$ was quoted at
$[_N$ \$ 366.50 $_N]$ $[_N$ an ounce $_N]$ . \\
tagged:\\
In/O early/I trading/I in/O Hong/I Kong/I 
Monday/B ,/O gold/I was/O quoted/O at/O
\$/I 366.50/I an/B ounce/I ./O 
\end{quote}

Other representations for NP chunking can be used as well.
An example is the representation used in \cite{ratnaparkhi98} 
where all the chunk-initial words receive the same
start tag  (analogous to the {\tt B} tag) while the remainder of the
words in the chunk are paired with a different tag.
This removes tagging ambiguities.
In the Ratnaparkhi representation equal noun phrases receive the
same tag sequence regardless of the context in which they appear.

The data representation choice might influence the performance of
chunking systems.
In this paper we discuss how large this influence is. 
Therefore we will compare seven different data representation formats 
for the baseNP recognition task.
We are particularly interested in finding out whether with one of the 
representation formats the best reported results for this task can
be improved.
The second section of this paper presents the general setup of the
experiments.
The results can be found in the third section.
In the fourth section we will describe some related work.

\begin{table*}[t]
\begin{center}
\begin{tabular}{|c|ccccccccccccccccc|}\hline
IOB1 &  O&I&I&O&I&I&B&O&I&O&O&O&I&I&B&I&O \\
IOB2 &  O&B&I&O&B&I&B&O&B&O&O&O&B&I&B&I&O \\
IOE1 &  O&I&I&O&I&E&I&O&I&O&O&O&I&E&I&I&O \\
IOE2 &  O&I&E&O&I&E&E&O&E&O&O&O&I&E&I&E&O \\
IO   &  O&I&I&O&I&I&I&O&I&O&O&O&I&I&I&I&O \\
$[$  &  .&$[$&.&.&$[$&.&$[$&.&$[$&.&.&.&$[$&.&$[$&.&. \\
$]$  &  .&.&$]$&.&.&$]$&$]$&.&$]$&.&.&.&.&$]$&.&$]$&. \\\hline
\end{tabular}
\end{center}
\caption{The chunk tag sequences for the example sentence
{\it In early trading in Hong Kong Monday , gold was quoted at \$
366.50 an ounce . } 
for seven different tagging formats.
The {\tt I} tag has been used for words inside a baseNP, {\tt O}
for words outside a baseNP, {\tt B} and {\tt [} for 
baseNP-initial words and {\tt E} and {\tt ]} for baseNP-final words.
}
\label{tab-formats}
\end{table*}

\section{Methods and experiments}

In this section we present and explain the data representation formats
and the machine learning algorithm that we have used.
In the final part we describe the feature representation used in
our experiments.

\subsection{Data representation}

We have compared four complete and three partial data representation
formats for the baseNP recognition task presented in \cite{ramshaw95}.
The four complete formats all use an {\tt I} tag for words that are
inside a baseNP and an {\tt O} tag for words that are outside a baseNP.
They differ in their treatment of chunk-initial and chunk-final words:

\begin{tabular}{cp{5.2cm}}
IOB1 & The first word inside a baseNP immediately following another
       baseNP receives a B tag \cite{ramshaw95}.\\
IOB2 & All baseNP-initial words receive a B tag \cite{ratnaparkhi98}.\\
IOE1 & The final word inside a baseNP immediately preceding another
       baseNP receives an E tag.\\
IOE2 & All baseNP-final words receive an E tag.
\end{tabular}

We wanted to compare these data representation formats with a 
standard bracket representation.
We have chosen to divide bracketing experiments in two parts: one for
recognizing opening brackets and one for recognizing closing brackets.
Additionally we have worked with another partial representation which
seemed promising: a tagging representation which disregards boundaries
between adjacent chunks.
These boundaries can be recovered by combining this format with one of
the bracketing formats.
Our three partial representations are:

\begin{tabular}{cp{5.5cm}}
[    & All baseNP-initial words receive an [ tag, other words receive a
       . tag.\\
]    & All baseNP-final words receive a ] tag, other words receive a
       . tag.\\
IO   & Words inside a baseNP receive an {\tt I} tag, others receive 
       an {\tt O} tag.
\end{tabular}

These partial representations can be combined in three pairs which 
encode the complete baseNP structure of the data:

\begin{tabular}{cp{5.1cm}}
[ + ] &    A word sequence is regarded as a baseNP if the first word 
           has received an [ tag, the final word has received a ]
           tag and these are the only brackets that have been assigned to
           words in the sequence.\\
$[$ + IO & In the IO format, tags of words that have received an {\tt I}
           tag and an [ tag are changed into {\tt B} tags.
           The result is interpreted as the IOB2 format.\\
IO + $]$ & In the IO format, tags of words that have received an {\tt I}
           tag and a ] tag are changed into {\tt E} tags.
           The result is interpreted as the IOE2 format.
\end{tabular}

\vspace*{0.3cm}

Examples of the four complete formats and the three partial formats
can be found in table \ref{tab-formats}.

\begin{table*}[t]
\begin{center}
\begin{tabular}{|c|c|c|}\cline{2-3}
\multicolumn{1}{c|}{}
& word/POS context & F$_{\beta=1}$ \\\hline
IOB1      &  L=2/R=1             & 89.17\\
IOB2      &  L=2/R=1             & 88.76\\
IOE1      &  L=1/R=2             & 88.67\\
IOE2      &  L=2/R=2             & 89.01\\
$[$ + $]$ &  L=2/R=1  +  L=0/R=2 & 89.32\\
$[$ + IO  &  L=2/R=0  +  L=1/R=1 & 89.43\\
IO + $]$  &  L=1/R=1  +  L=0/R=2 & 89.42\\\hline
\end{tabular}
\end{center}
\caption{
Results first experiment series:
the best F$_{\beta=1}$ scores for different left (L) and right
(R) word/POS tag pair context sizes for the seven representation
formats using 5-fold cross-validation on section 15 of the WSJ corpus.}
\label{tab-result1}
\end{table*}
 
\subsection{Memory-Based Learning}

We have build a baseNP recognizer by training a machine learning
algorithm with correct tagged data and testing it with unseen data.
The machine learning algorithm we used was a Memory-Based Learning
algorithm (MBL).
During training it stores a symbolic feature representation of a word 
in the training data together with its classification (chunk tag).
In the testing phase the algorithm compares a feature
representation of a test word with every training data item and
chooses the classification of the training item which is closest 
to the test item.

In the version of the algorithm that we have used, {\sc ib1-ig},
the distances between feature representations are computed as
the weighted sum of distances between individual features
\cite{timbl98}.
Equal features are defined to have distance 0, while the distance
between other pairs is some feature-dependent value.
This value is equal to the information gain of the feature,
an information theoretic measure which contains the normalized 
entropy decrease of the classification set caused by the presence
of the feature.
Details of the algorithm can be found in 
\cite{timbl98}\footnote{{\sc ib1-ig} is a part of the TiMBL software
package which is available from http://ilk.kub.nl
}.

\subsection{Representing words with features}

An important decision in an MBL experiment is the
choice of the features that will be used for representing the data.
{\sc ib1-ig} is thought to be less sensitive to redundant
features because of the data-dependent feature weighting that is
included in the algorithm.
We have found that the presence of redundant features has a 
negative influence on the performance of the baseNP recognizer.

In \cite{ramshaw95} a set of transformational rules is used for 
modifying the classification of words.
The rules use context information of the words, the part-of-speech
tags that have been assigned to them and the chunk tags that are
associated with them.
We will use the same information as in our feature representation for
words.  

In TBL, rules with different context information
are used successively for solving different problems.
We will use the same context information for all data.
The optimal context size will be determined by comparing the results
of different context sizes on the training data.
Here we will perform four steps.
We will start with testing different context sizes of words with their
part-of-speech tag.
After this, we will use the classification results of the best context
size for determining the optimal context size for the classification
tags. 
As a third step, we will evaluate combinations of classification
results and find the best combination.
Finally we will examine the influence of an MBL algorithm parameter:
the number of examined nearest neighbors. 

\begin{table*}
\begin{center}
\begin{tabular}{|c|c|c|c|}\cline{2-4}
\multicolumn{1}{c|}{}
          & word/POS context    & chunk tag context & F$_{\beta=1}$\\\hline
IOB1      & L=2/R=1             & 1/2               & 90.12\\
IOB2      & L=2/R=1             & 1/0               & 89.30\\
IOE1      & L=1/R=2             & 1/2               & 89.55\\
IOE2      & L=1/R=2             & 0/1               & 89.73\\
$[$ + $]$ & L=2/R=1  +  L=0/R=2 & 0/0 + 0/0         & 89.32\\
$[$ + IO  & L=2/R=0  +  L=1/R=1 & 0/0 + 1/1         & 89.78\\
IO + $]$  & L=1/R=1  +  L=0/R=2 & 1/1 + 0/0         & 89.86\\\hline
\end{tabular}
\end{center}
\caption{
Results second experiment series:
the best F$_{\beta=1}$ scores for different left (L) and right
(R) chunk tag context sizes for the seven representation
formats using 5-fold cross-validation on section 15 of the WSJ corpus.
}
\label{tab-result2}
\end{table*}

\begin{table*}
\begin{center}
\begin{tabular}{|c|c|c|c|c|}\cline{2-5}
\multicolumn{1}{c|}{}
          & word/POS    & chunk tag & combinations    & F$_{\beta=1}$ \\\hline
IOB1      & 2/1         & 1/1       & 0/0 1/1 2/2 3/3     & 90.53\\
IOB2      & 2/1         & 1/0       & 2/1                 & 89.30\\
IOE1      & 1/2         & 1/2       & 0/0 1/1 2/2 3/3     & 90.03\\
IOE2      & 1/2         & 0/1       & 1/2                 & 89.73\\
$[$ + $]$ & 2/1 + 0/2   & 0/0 + 0/0 & - + -               & 89.32\\
$[$ + IO  & 2/0 + 1/1   & 0/0 + 1/1 & - + 0/1 1/2 2/3 3/4 & 89.91\\
IO + $]$  & 1/1 + 0/2   & 1/1 + 0/0 & 0/1 1/2 2/3 3/4 + - & 90.03\\\hline
\end{tabular}
\end{center}
\caption{
Results third experiment series:
the best F$_{\beta=1}$ scores for different combinations of chunk tag
context sizes for the seven representation formats using 5-fold
cross-validation on section 15 of the WSJ corpus. 
}
\label{tab-result3}
\end{table*}

\section{Results}

We have used the baseNP data presented in
\cite{ramshaw95}\footnote{The data described in \cite{ramshaw95} is
available from ftp://ftp.cis.upenn.edu/pub/chunker/}.
This data was divided in two parts.
The first part was training data and consisted of 211727 words taken
from sections 15, 16, 17 and 18 from the Wall Street Journal corpus
(WSJ). 
The second part was test data and consisted of 47377 words taken from
section 20 of the same corpus.
The words were part-of-speech (POS) tagged with the Brill tagger and
each word was classified as being inside or outside a baseNP
with the IOB1 representation scheme. 
The chunking classification was made by \cite{ramshaw95} based on the
parsing information in the WSJ corpus.

The performance of the baseNP recognizer can be measured in
different ways:
by computing 
the percentage of correct classification tags (accuracy),
the percentage of recognized baseNPs that are correct (precision) and
the percentage of baseNPs in the corpus that are found (recall).
We will follow \cite{argamon98} and use a combination of 
the precision and recall rates: 
F$_{\beta=1}$ = (2*precision*recall)/(precision+recall).

In our first experiment series we have tried to discover the best 
word/part-of-speech tag context for each representation format.
For computational reasons we have limited ourselves to working 
with section 15 of the WSJ corpus.
This section contains 50442 words.
We have run 5-fold cross-validation experiments with all combinations
of left and right contexts of word/POS tag pairs in the size range 
0 to 4. 
A summary of the results can be found in table \ref{tab-result1}.

The baseNP recognizer performed best with relatively small word/POS
tag pair contexts.
Different representation formats required different context sizes for
optimal performance.
All formats with explicit open bracket information preferred larger
left context and most formats with explicit closing bracket
information preferred larger right context size.
The three combinations of partial representations systematically
outperformed the four complete representations.
This is probably caused by the fact that they are able to use two
different context sizes for solving two different parts of the 
recognition problem.

In a second series of experiments we used a "cascaded" classifier.
This classifier has two stages (cascades). 
The first cascade is similar to the classifier described in the
first experiment. 
For the second cascade we added the classifications
of the first cascade as extra features.
The extra features consisted of the left and the right context of
the classification tags.
The focus chunk tag (the classification of the current word)
accounts for the correct classification in about 95\% of the cases.
The MBL algorithm assigns a
large weight to this input feature and this makes it harder for the
other features to contribute to a good result.
To avoid this we have refrained from using this tag.
Our goal was to find out the optimal number of extra classification
tags in the input.
We performed 5-fold cross-validation experiments with all combinations
of left and right classification tag contexts in the range 0 tags to 
3 tags. 
A summary of the results can be found in table 
\ref{tab-result2}\footnote{In a number of cases a different base
configuration in one experiment series outperformed the best base
configuration found in the previous series.
In the second series L/R=1/2 outperformed 2/2 for IOE2 when chunk
tags were added and in the third series chunk tag context 1/1
outperformed 1/2 for IOB1 when different combinations were tested.}.
We achieved higher F$_{\beta=1}$ for all representations except
for the bracket pair representation.

The third experiment series was similar to the second but instead of
adding output of one experiment we added classification results of
three, four or five experiments of the first series.
By doing this we supplied the learning algorithm with
information about different context sizes.
This information is available to TBL in the rules
which use different contexts.
We have limited ourselves to examining all successive
combinations of three, four and five experiments of the lists
(L=0/R=0, 1/1, 2/2, 3/3, 4/4), (0/1, 1/2, 2/3, 3/4) and 
(1/0, 2/1, 3/2, 4/3).
A summary of the results can be found in table \ref{tab-result3}.
The results for four representation formats improved.

\begin{table*}
\begin{center}
\begin{tabular}{|c|c|c|c|c|}\cline{2-5}
\multicolumn{1}{c|}{}
     & word/POS  & chunk tag & combinations          & F$_{\beta=1}$ \\\hline
IOB1 & 3/3(k=3)  & 1/1 & 0/0(1) 1/1(1) 2/2(3) 3/3(3) & 90.89 $\pm$ 0.63\\
IOB2 & 3/3(k=3)  & 1/0 & 3/3(3)                      & 89.72 $\pm$ 0.79\\
IOE1 & 2/3(k=3)  & 1/2 & 0/0(1) 1/1(1) 2/2(3) 3/3(3) & 90.12 $\pm$ 0.27\\
IOE2 & 2/3(k=3)  & 0/1 & 2/3(3)                      & 90.02 $\pm$ 0.48\\
$[$ + $]$ & 4/3(3) + 4/4(3) & 0/0 + 0/0 & - + -      & 90.08 $\pm$ 0.57\\
$[$ + IO  & 4/3(3) + 3/3(3) & 0/0 + 1/1 & - + 0/1(1) 1/2(3) 2/3(3) 3/4(3) & 
                                                       90.35 $\pm$ 0.75\\
IO + $]$  & 3/3(3) + 2/3(3) & 1/1 + 0/0 & 0/1(1) 1/2(3) 2/3(3) 3/4(3) + - &
                                                       90.23 $\pm$ 0.73\\\hline
\end{tabular}
\end{center}
\caption{
Results fourth experiment series:
the best F$_{\beta=1}$ scores for different combinations of left
and right classification tag context sizes for the seven
representation formats using 5-fold cross-validation on section 15 of
the WSJ corpus obtained with {\sc ib1-ig} parameter {\tt k}=3.
IOB1 is the best representation format but the differences with the
results of the other formats are not significant.
}
\label{tab-result4}
\end{table*}

In the fourth experiment series we have experimented with a different
value for the number of nearest neighbors examined by the {\sc ib1-ig}
algorithm (parameter {\tt k}).
This algorithm standardly uses the single training item closest to the 
test item.
However \cite{daelemans99} report that for baseNP recognition better
results can be obtained by making the algorithm consider the
classification values of the three closest training items.
We have tested this by repeating the first experiment series and part
of the third experiment series for {\tt k}=3.
In this revised version we have repeated the best experiment of the 
third series with the results for {\tt k}=1 replaced by the {\tt k}=3
results whenever the latter outperformed the first in the revised
first experiment series.
The results can be found in table \ref{tab-result4}.
All formats benefited from this step.
In this final experiment series the best results were obtained with
IOB1 but the differences with the results of the other formats are
not significant.

We have used the optimal experiment configurations that we had
obtained from the fourth experiment series for processing the complete 
\cite{ramshaw95} data set.
The results can be found in table \ref{tab-result-all}.
They are better than the results for section 15 because more training
data was used in these experiments.
Again the best result was obtained with IOB1 (F$_{\beta=1}$=92.37)
which is an improvement of the
best reported F$_{\beta=1}$ rate for this data set 
(\cite{ramshaw95}: 92.03).

\begin{table*}
\begin{center}
\begin{tabular}{|cc|c|c|c|c|}\cline{3-6}
\multicolumn{2}{c|}{}
          & accuracy       & precision & recall    & F$_{\beta=1}$ \\\hline
&IOB1      & 97.58\%        & 92.50\%   & 92.25\%   & 92.37\\
&IOB2      & 96.50\%        & 91.24\%   & 92.32\%   & 91.78\\
&IOE1      & 97.58\%        & 92.41\%   & 92.04\%   & 92.23\\
&IOE2      & 96.77\%        & 91.93\%   & 92.46\%   & 92.20\\
&$[$ + $]$ & -              & 93.66\%   & 90.81\%   & 92.22\\
&$[$ + IO  & -              & 91.47\%   & 92.61\%   & 92.04\\
&IO + $]$  & -              & 91.25\%   & 92.54\%   & 91.89\\\hline
\multicolumn{2}{|c|}{\cite{ramshaw95}} & 97.37\% & 91.80\% & 92.27\% & 92.03\\
\multicolumn{2}{|c|}{\cite{veenstra98}} & 97.2\% & 89.0\% & 94.3\% & 91.6 \\
\multicolumn{2}{|c|}{\cite{argamon98}} & -       & 91.6 \% & 91.6\% & 91.6 \\
\multicolumn{2}{|c|}{\cite{cardie98}}  & -       & 90.7\% & 91.1\% & 90.9 \\\hline
\end{tabular}
\end{center}
\caption{
The F$_{\beta=1}$ scores for the \cite{ramshaw95} test set after
training with their training data set.
The data was processed with the optimal input feature combinations
found in the fourth experiment series. 
The accuracy rate contains the fraction of chunk tags that was correct.
The other three rates regard baseNP recognition.
The bottom part of the table shows some other reported results with
this data set. 
With all but two formats {\sc ib1-ig} achieves better F$_{\beta=1}$ rates
than the best published result in \cite{ramshaw95}.
}
\label{tab-result-all}
\end{table*}

We would like to apply our learning approach to the large data
set mentioned in \cite{ramshaw95}: Wall Street Journal corpus sections
2-21 as training material and section 0 as test material.
With our present hardware applying our optimal experiment
configuration to this data would require several months of computer
time.
Therefore we have only used the best stage 1 approach with IOB1 tags:
a left and right context of three words and three POS tags combined
with k=3.
This time the chunker achieved a F$_{\beta=1}$ score of 93.81 which is half
a point better than the results obtained by \cite{ramshaw95}: 93.3
(other chunker rates for this data:
accuracy: 98.04\%; precision: 93.71\%; recall: 93.90\%).

\section{Related work}

The concept of chunking was introduced by Abney in
\cite{abney91}. 
He suggested to develop a chunking parser which uses a two-part
syntactic analysis: creating word chunks (partial trees) and
attaching the chunks to create complete syntactic trees.
Abney obtained support for such a chunking stage from psycholinguistic
literature.

Ramshaw and Marcus used transformation-based learning (TBL) for
developing two chunkers \cite{ramshaw95}.
One was trained to recognize baseNPs and the other was trained to
recognize both NP chunks and VP chunks.
Ramshaw and Marcus approached the chunking task as a tagging problem.
Their baseNP  training and test data from the Wall Street Journal
corpus are still being used as benchmark data for current chunking
experiments. 
\cite{ramshaw95} shows that baseNP recognition
(F$_{\beta=1}$=92.0) is easier than finding both NP and VP chunks
(F$_{\beta=1}$=88.1) 
and that increasing the size of the training data increases the
performance on the test set.

The work by Ramshaw and Marcus has inspired three other groups to
build chunking algorithms.
\cite{argamon98} introduce Memory-Based Sequence Learning and use
it for different chunking experiments.
Their algorithm stores sequences of POS tags with chunk brackets 
and uses this information for recognizing chunks in unseen data.
It performed slightly worse on baseNP recognition than the
\cite{ramshaw95} experiments (F$_{\beta=1}$=91.6).
\cite{cardie98} uses a related method but they only store 
POS tag sequences forming complete baseNPs.
These sequences were applied to unseen tagged data after which 
post-processing repair rules were used for fixing some frequent errors.
This approach performs worse than other reported approaches
(F$_{\beta=1}$=90.9).

\cite{veenstra98} uses cascaded decision tree learning (IGTree) for 
baseNP recognition.
This algorithm stores context information of words, POS tags and
chunking tags in a decision tree and 
classifies new items by comparing them to the training items.
The algorithm is very fast and
it reaches the same performance as \cite{argamon98}
(F$_{\beta=1}$=91.6). 
\cite{daelemans99} uses cascaded MBL ({\sc ib1-ig})
in a similar way for several tasks among which baseNP recognition.
They do not report F$_{\beta=1}$ rates but their tag accuracy
rates are a lot better than accuracy rates reported by others.
However, they use the \cite{ramshaw95} data set in a different 
training-test division (10-fold cross validation) which makes it
difficult to compare their results with others.

\section{Concluding remarks}

We have compared seven different data formats for the recognition
of baseNPs with memory-based learning ({\sc ib1-ig}).
The IOB1 format, introduced in \cite{ramshaw95}, consistently came
out as the best format.
However, the differences with other formats were not significant. 
Some representation formats achieved better precision rates, others
better recall rates.
This information is useful for tasks that require chunking structures
because some tasks might be more interested in high precision rates
while others might be more interested in high recall rates.

The {\sc ib1-ig} algorithm has been able to improve the best reported
F$_{\beta=1}$ rates for a standard data set (92.37 versus 
\cite{ramshaw95}'s 92.03).
This result was aided by using non-standard parameter values
(k=3) and the algorithm was sensitive for redundant input features.
This means that finding an optimal performance or this task
requires searching a large parameter/feature configuration space.
An interesting topic for future research would be to embed {\sc ib1-ig}
in a standard search algorithm, like hill-climbing, and explore this
parameter space.
Some more room for improved performance lies in computing the POS tags
in the data with a better tagger than presently used.

\bibliographystyle{acl}

\end{document}